# Sign Language Recognition System using TensorFlow Object Detection API

Sharvani Srivastava, Amisha Gangwar, Richa Mishra, Sudhakar Singh*[0000-0002-0710-924X]

Department of Electronics and Communication, University of Allahabad, Prayagraj, India
```
{sharvanisri28, amishagangwar21}@gmail.com, {richa_mishra,
                sudhakar}@allduniv.ac.in
```
*Corresponding Author

**Abstract.** Communication is defined as the act of sharing or exchanging information, ideas or feelings. To establish communication between two people, both of them are required to have knowledge and understanding of a common language. But in the case of deaf and dumb people, the means of communication are different. Deaf is the inability to hear and dumb is the inability to speak. They communicate using sign language among themselves and with normal people but normal people do not take seriously the importance of sign language. Not everyone possesses the knowledge and understanding of sign language which makes communication difficult between a normal person and a deaf and dumb person. To overcome this barrier, one can build a model based on machine learning. A model can be trained to recognize different gestures of sign language and translate them into English. This will help a lot of people in communicating and conversing with deaf and dumb people. The existing Indian Sing Language Recognition systems are designed using machine learning algorithms with single and double-handed gestures but they are not real-time. In this paper, we propose a method to create an Indian Sign Language dataset using a webcam and then using transfer learning, train a TensorFlow model to create a real-time Sign Language Recognition system. The system achieves a good level of accuracy even with a limited size dataset.

## 1 Introduction

Communication can be defined as the act of transferring information from one place, person, or group to another. It consists of three components: the speaker, the message that is being communicated, and the listener. It can be considered successful only when whatever message the speaker is trying to convey is received and understood by the listener. It can be divided into different categories as follows [1]: formal and informal communication, oral (face-to-face and distance) and written communication, non-verbal, grapevine, feedback, and visual communication, and the active listening. The formal communication (official communication) is steered through the channels that are pre-determined. The unofficial or grapevine communication is the spontaneous communication between individuals in one's profession that does not have any formal protocol or structure. The oral communication (face-to-face and distance) is the communication in which words are exchanged between people who are present in front or at a distance (with the help of technology including voice and video calls, webinars, etc.). The written communication is the communication in which letters, emails, notices, or any other written form is used for communicating. The non-verbal communication is the communication that uses gestures, facial expressions, body language, etc. The feedback communication happens when a person gives feedback on some product or service provided by an individual or a company. The visual communication occurs when a person gets information from a visual source like televisions, social networking, or any other source. Active listening is when a person listens to and understands what the other individual is trying to convey so that the communication becomes more meaningful and effective [1].

Non-verbal communication helps deaf and dumb people to communicate amongst themselves and with others. Deaf is a disability that impairs a person's hearing ability and makes them incapable to hear while dumb is a disability that impairs the speaking ability and makes them incapable to speak. Not being able to speak or listen makes it difficult to establish communication with others. This is where sign languages come into the role, it enables a person to communicate without words. But a problem still exists, not many people possess the knowledge of sign language. Deaf and dumb may be able to communicate amongst themselves using sign languages but it is still difficult for them to communicate with people having normal hearing and vice-versa due to the lack of knowledge of sign languages. This issue can be resolved by the use of a technology-driven solution. By using such a solution, one can easily translate the gestures of sign language into the commonly spoken language, English.

A lot of research has been done in this field and there is still a need for further research. For gesture translation, data gloves, motion capturing systems, or sensors have been used [2]. Vision-based SLR systems have also been developed previously [3]. The existing Indian Sign Language Recognition system was developed using machine learning algorithms with MATLAB [4]. Authors have worked on single-handed and double-handed gestures. They used two algorithms to train their system, K Nearest Neighbours Algorithm and Back Propagation Algorithm. Their system achieved 93-96% accuracy. Though being highly accurate, it is not a real-time SLR system. The objective of this paper is to develop a real-time SLR system using TensorFlow object detection API and train it using a dataset that will be created using a webcam.

The rest of this paper after the introduction is organized as follows. Section 2 presents the related work on the SLR system. Section 3 describes the data acquisition and generation. Section 4 focuses on the methodology of the developed system. Section 5 presents the experimental evaluation of the system, and finally, Section 6 concludes the paper with future work.

## 2  Related work

Sign languages are defined as an organized collection of hand gestures having specific meanings which are employed from the hearing impaired people to communicate in everyday life [3]. Being visual languages, they use the movements of hands, face, and body as communication mediums. There are over 300 different sign languages available all around the world [5]. Though there are so many different sign languages, the percentage of population knowing any of them is low which makes it difficult for the specially-abled people to communicate freely with everyone. SLR provides a means to communicate in sign language without knowing it. It recognizes a gesture and translates it into a commonly spoken language like English.

SLR is a very vast topic for research where a lot of work has been done but still various things need to be addressed. The machine learning techniques allow the electronic systems to take decisions based on experience i.e. data. The classification algorithms need two datasets – training dataset and testing dataset. The training set provides experiences to the classifier and the model is tested using the testing set [6]. Many authors have developed efficient data acquisition and classification methods [3][7]. Based on data acquisition method, previous work can be categorized into two approaches: the direct measurement methods and the vision-based approaches [3]. The direct measurement methods are based on motion data gloves, motion capturing systems, or sensors. The motion data extracted can supply accurate tracking of fingers, hands, and other body parts which leads to robust SLR methodologies development. The vision-based SLR approaches rely on the extraction of discriminative spatial and temporal from RGB images. Most of the vision-based methods initially try to track and extract the hand regions before their classification to gestures [3]. Hand detection is achieved by semantic segmentation and skin colour detection as the skin colour is usually distinguishable easily [8][9]. Though, because the other body parts like face and arms can be mistakenly recognized as hands, so, the recent hand detection methods also use the face detection and subtraction, and background subtraction to recognize only the moving parts in a scene [10][11]. To attain accurate and robust hands tracking, particularly in cases of obstructions, authors employed filtering techniques, for example, Kalman and particle filters [10][12].

For data acquisition by either the direct measurement or the vision-based approaches, different devices need to be used. The primary device employed as input process in SLR system is camera [13]. There are other devices available that are used for input such as Microsoft Kinect which provides colour video stream and depth video stream all together. The depth data helps in background segmentation. Apart from the devices, other methods used for acquiring data are accelerometer and sensory gloves. Another system that is used for data acquisition is Leap Motion Controller (LMC) [14][15] – it is a touchless controller developed by technology company "Leap Motion" now called "Ultraleap" based in San Francisco. Approximately, it can operate around 200 frames per second and can detect and track the hands, fingers, and objects that look alike fingers. Most of the researchers collect their training dataset by recording it from their signer as finding a sign language dataset is a problem [2].

Different processing methods have been used for creating an SLR system [16][17][18]. Hidden Markov Model (HMM) has been widely used in SLR [12]. The various HMM that have been used are Multi Stream HMM (MSHMM) which is based on the two standard single-stream HMMs, Light-HMM, and Tied-Mixture Density-HMM [2]. The other processing models that have been used are neural network [19][20][21][22][23], ANN [24], Naïve Bayes Classifier (NBC), and Multilayer Perceptron (MLP) [14], unsupervised neural network Self-Organizing Map (SOM) [25], Self-Organizing Feature Map (SOFM), Simple Recurrent Network (SRN) [26], Support Vector Machine (SVM) [27], 3D convolutional residual network [28]. Researchers have also used self-designed methods like the wavelet-based method [29] and Eigen Value Euclidean Distance [30].

The use of different processing methods or application systems has given different accuracy results. The Light-HMM gave 83.6% accuracy result, the MSHMM gave 86.7% accuracy result, SVM gave 97.5% accuracy result, Eigen Value gave 97% accuracy result, Wavelet Family gave 100% accuracy result [2][31][22][32]. Though different models have given high accuracy results, but the accuracy does not depend only on the processing model used, it depends upon various factors such as size of the dataset, clarity of images of the dataset depending upon data acquisition methods, devices used, etc.

There are two types of SLR systems – isolated SLR and continuous SLR. In isolated SLR, the system is trained to recognize a single gesture. Each image is labelled to represent an alphabet, a digit, or some special gesture. Continuous SLR is different from isolated gesture classification. In continuous SLR, the system is able to recognize and translate whole sentences instead of a single gesture [33][34].

Even with all the research that has been done in SLR, many inadequacies need to be dealt with by further research. Some of the issues and challenges that need to be worked on are as follows [33][2][4][6].

- Isolated SLR methods need to do strenuous labeling for each word.
- Continuous SLR methods make use of isolated SLR systems as building blocks with temporal segmentation as pre-processing, which is non-trivial and unescapably proliferates errors into subsequent steps, and sentence synthesis as post-processing.
- Devices needed for data acquisition are costly, a cheap method is needed for SLR systems to be commercialized.
- Web camera is an alternative to higher specification camera but the image is blurred so, the quality is compromised.
- Data acquisition by sensors also has some issues e.g., noise, bad human manipulation, bad ground connection, etc.
- Vision-based methodologies introduce inaccuracies due to overlapping of hand and finger.
- Large datasets are not available.
- There are misconceptions about sign languages like sign language is same around the world, while sign language is based upon the spoken language.

- Indian Sign Language is communicated using hand gestures made by a single hand and double hands due to which there are two types of gestures representing the same thing.

In this paper, the dataset that will be used is created using Python and OpenCV with the help of a webcam. The SLR system that is being developed is a real-time detection system.

## 3   Data acquisition

A real-time sign language detection system is being developed for Indian Sign Language. For data acquisition, images are captured by webcam using Python and OpenCV. OpenCV provides functions which are primarily aimed at the real-time computer vision. It accelerates the use of machine perception in commercial products and provides a common infrastructure for the computer vision-based applications. The OpenCV library has more than 2500 efficient computer vision and machine learning algorithms which can be used for face detection and recognition, object identification, classification of human actions, tracking camera and object movements, extracting 3D object models, and many more [35].

The created dataset is made up of signs representing alphabets in Indian Sign Language [36] as shown in Fig. 1. For every alphabet, 25 images are captured to make the dataset. The images are captured in every 2 seconds providing time to record gesture with a bit of difference every time and a break of five seconds are given between two individual signs, i.e., to change the sign of one alphabet to the sign of a different alphabet, five seconds interval is provided. The captured images are stored in their respective folder.

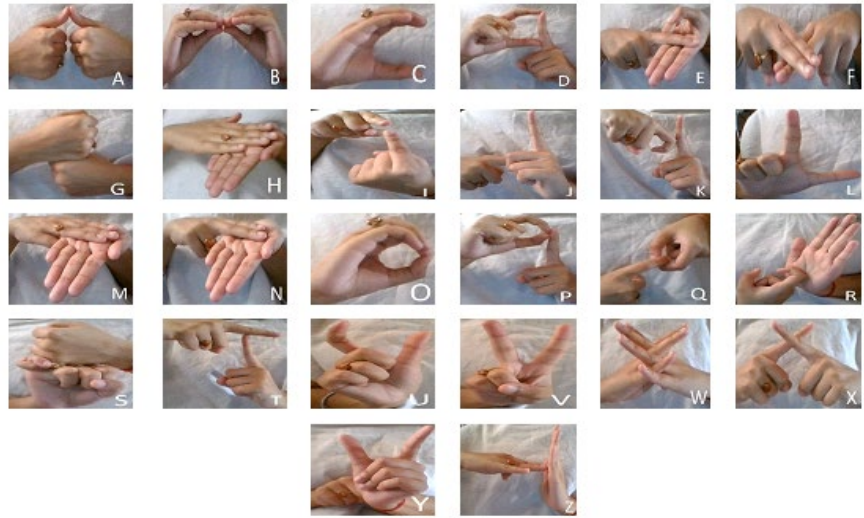

**Fig. 1.** Indian Sign Language Alphabets

For data acquisition, dependencies like cv2, i.e., OpenCV, os, time, and uuid have been imported. The dependency os is used to help work with file paths. It comes under standard utility modules of Python and provides functions for interacting with the operating systems. With the help of the time module in Python, time can be represented in multiple ways in code like objects, numbers, and strings. Apart from representing time, it can be used to measure code efficiency or wait during code execution. Here, it is used to add breaks between the image capturing in order to provide time for hand movements. The uuid library is used in naming the image files. It helps in the generation of random objects of 128 bits as ids providing uniqueness as the ids are generated on the basis of time and computer hardware.

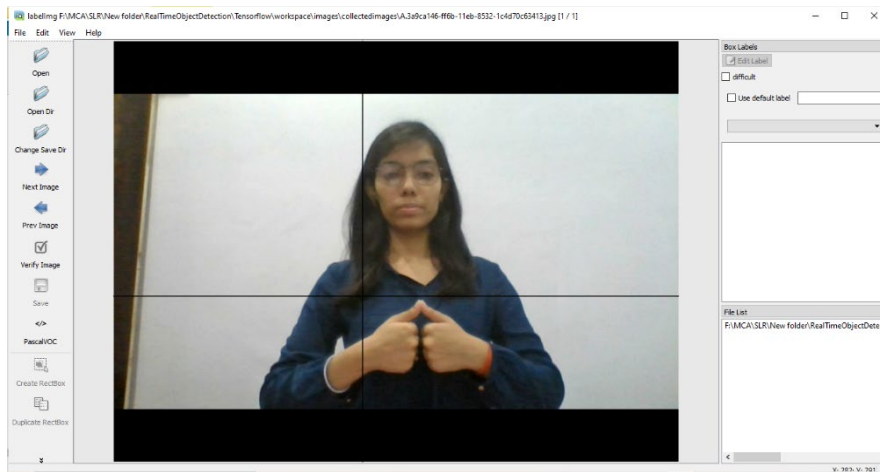

**Fig. 2.** Selecting a portion of the image to label it

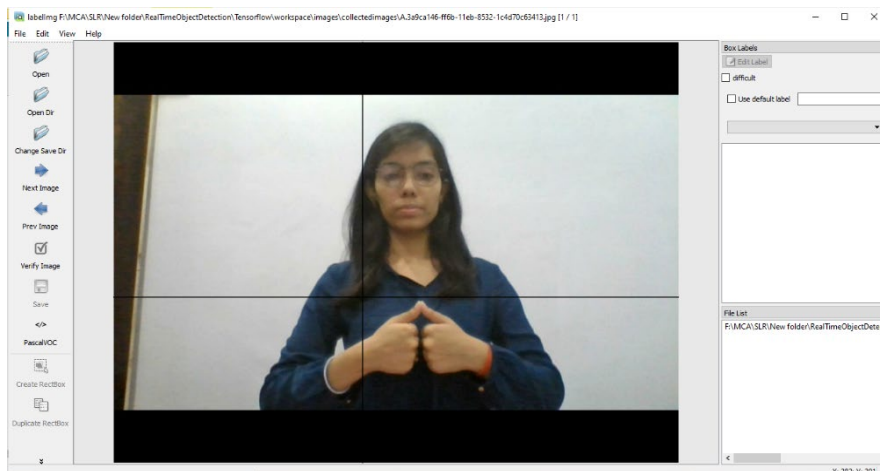

**Fig. 3.** Labelling the selected portion

Once all the images have been captured, they are then one by one labelled using the LabelImg package. LabelImg is a free open-source tool for graphically labelling images. The hand gesture portion of the image is labelled by what the gesture in the box or the sign represents as shown in Fig. 2 and Fig. 3. On saving the labelled image, its XML file is created. The XML files have all the details of the images including the detail of the labelled portion. After labelling all the images, their XML files are available. This is used for creating the TF (TensorFlow) records. All the images along with their XML files are then divided into training data and validation data in the ratio of 80:20. From 25 images of an alphabet, 20 (80%) of them were taken and stored as a training dataset and the remaining 5 (20%) were taken and stored as validation dataset. This task was performed for all the images of all 26 alphabets.

## 4  Methodology

The proposed system is designed to develop a real-time sign language detector using a TensorFlow object detection API and train it through transfer learning for the created dataset [37]. For data acquisition, images are captured by a webcam using Python and OpenCV following the procedure described under Section 3.

Following the data acquisition, a labeled map is created which is a representation of all the objects within the model, i.e., it contains the label of each sign (alphabet) along with their id. The label map contains 26 labels, each one representing an alphabet. Each label has been assigned a unique id ranging from 1 to 26. This will be used as a reference to look up the class name. TF records of the training data and the testing data are then created using generate_tfrecord which is used to train the TensorFlow object detection API. TF record is the binary storage format of TensorFlow. Binary files usage for storage of the data significantly impacts the performance of the import pipeline consequently, the training time of the model. It takes less space on a disk, copies fast, and can efficiently be read from the disk.

The open-source framework, TensorFlow object detection API makes it easy to develop, train and deploy an object detection model. They have their framework called the TensorFlow detection model zoo which offers various models for detection that have been pre-trained on the COCO 2017 dataset. The pre-trained TensorFlow model that is being used is SSD MobileNet v2 320x320. The SSD MobileNet v2 Object detection model is combined with the FPN-lite feature extractor, shared box predictor, and focal loss with training images scaled to 320x320. Pipeline configuration, i.e., the configuration of the pre-trained model is set up and then updated for transfer learning to train it by the created dataset. For configuration, dependencies like TensorFlow, config_util, pipeline_pb2, and text_format have been imported. The major update that has been done is to change the number of classes which is initially 90 to 26, the number of signs (alphabets) that the model will be trained on. After setting up and updating the configuration, the model was trained in 10000 steps. The hyper-parameter used during the training was to set up the number of steps in which the model will be trained which was set up to 10000 steps. During the training, the model has some losses as classification loss, regularization loss, and localization loss. The localization loss is mismatched between the predicted bounding box correction and the true values. The formula of the localization loss [38] is given in Eq. (1) – (5).

$$L_{loc}(x, l, g) = \sum_{i \in Pos}^{N} \sum_{m \in \{cx, cy, w, h\}} x_{ij}^{k} smooth_{L1}(l_i^m - \hat{g}_j^m) \tag{1}$$

$$\hat{g}_j^{cx} = (g_j^{cx} - d_i^{cx})/d_i^w \tag{2}$$

$$\hat{g}_j^{cy} = (g_j^{cy} - d_i^{cy})/d_i^h \tag{3}$$

$$\hat{g}_j^w = \log(g_j^w/d_i^w) \tag{4}$$

$$\hat{g}_j^h = \log(g_j^h/d_i^h) \tag{5}$$

where, $N$ is the number of the matched default boxes, $l$ is the predicted bounding box, $g$ is the ground truth bounding box, $\hat{g}$ is the encoded ground truth bounding box and $x_{ij}^k$ is the matching indicator between default box $i$ and ground truth box $j$ of category $k$.

The classification loss is defined as the softmax loss over multiple classes. The formula of the classification loss [38] is as Eq. (6).

$$L_{conf}(x, c) = -\sum_{i \in Pos}^{N} x_{ij}^p \log(\hat{c}_i^p) - \sum_{i \in Neg} \log(\hat{c}_i^0) \tag{6}$$

where, $\hat{c}_i^p = \exp(c_i^p)/\sum_p \exp(c_i^p)$ is the softmax activated class score for default box $i$ with category $p$, $x_{ij}^p$ is the matching indicator between default box $i$ and the ground truth box $j$ of category $p$.

The different losses incurred during the experimentation are mentioned in the subsequent section. After training, the model is loaded from the latest checkpoint which makes it ready for real-time detection. After setting up and updating the configuration, the model will be ready for training. The trained model is loaded from the latest checkpoint which is created during the training of the model. This completes the model making it ready for real-time sign language detection.

The real-time detection is done using OpenCV and webcam again. For, real-time detection, cv2, and NumPy dependencies are used. The system detects signs in real-time and translates what each gesture means into English as shown in Fig. 5. The system is tested in real-time by creating and showing it different signs. The confidence rate of each sign (alphabet), i.e., how confident the system is in recognizing a sign (alphabet) is checked, noted, and tabulated for the result.

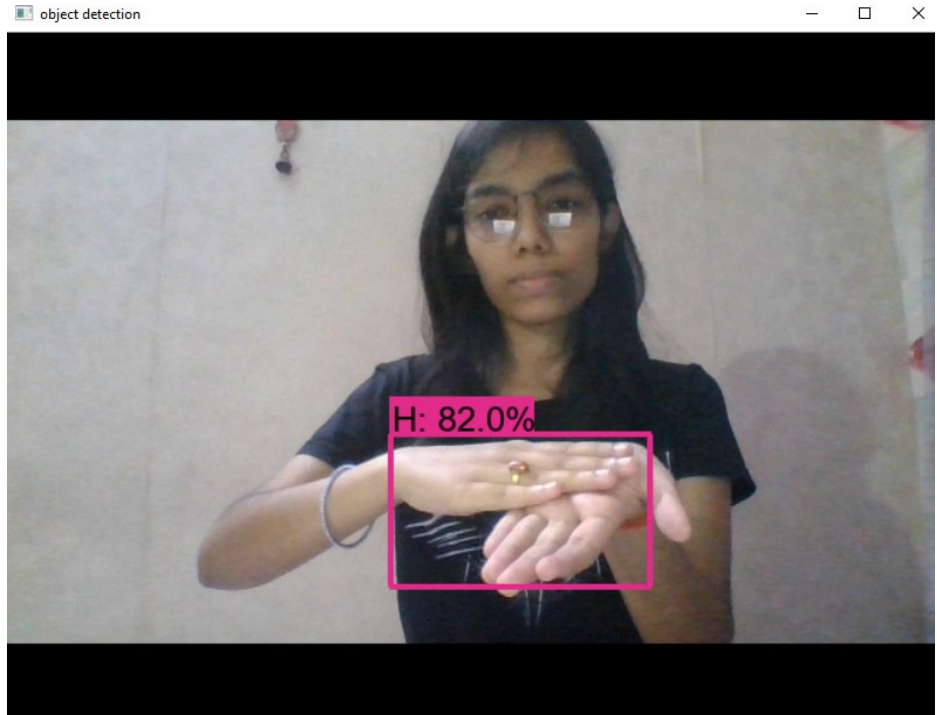

**Fig. 4.** Real-Time Sign Language Detection

## 5      Experimental Evaluation

### 5.1     Dataset and Experimental Setup

The dataset is created for Indian Sign Language where signs are alphabets of the English language. The dataset is created following the data acquisition method described in Section 3.

The experimentation was carried out on a system with an Intel i5 7$^{th}$ generation 2.70 GHz processor, 8 GB memory and webcam (HP TrueVision HD camera with 0.31 MP and 640x480 resolution), running Windows 10 operating system. The programming environment includes Python (version 3.7.3), Jupyter Notebook, OpenCV (version 4.2.0), TensorFlow Object Detection API.

### 5.2     Results and Discussion

The developed system is able to detect Indian Sign Language alphabets in real-time. The system has been created using TensorFlow object detection API. The pre-trained model that has been taken from the TensorFlow model zoo is SSD MobileNet v2 320x320. It has been trained using transfer learning on the created dataset which contains 650 images in total, 25 images for each alphabet.

The total loss incurred during the last part of the training, at 10,000 steps was 0.25, localization loss was 0.18, classification loss was 0.13, and regularization loss was 0.10 as shown in Fig. 4. Fig. 4 also shows that the lowest lost 0.17 was suffered at steps 9900.

**Fig. 5.** Loss incurred at different steps

The result of the system is based on the confidence rate and the average confidence rate of the system is 85.45%. For each alphabet, the confidence rate is recorded and tabulated in the result as shown in Table 1. The confidence rate of the system can be increased by increasing the size of the dataset which will boost up the recognition ability of the system. Thus, improving the result of the system and enhancing it.

**Table 1.** Confidence rate of each alphabet

| A | B | C | D | E | F | G | H | I |
|---|---|---|---|---|---|---|---|---|
| 94% | 98% | 90% | 90% | 70% | 96% | 73% | 97% | 95% |
| **J** | **K** | **L** | **M** | **N** | **O** | **P** | **Q** | **R** |
| 57% | 87% | 93% | 91% | 55% | 78% | 95% | 95% | 83% |
| **S** | **T** | **U** | **V** | **W** | **X** | **Y** | **Z** | |
| 86% | 81% | 87% | 86% | 87% | 88% | 90% | 80% | |

The state-of-the-art method of the Indian Sign Language Recognition system achieved 93-96% accuracy [4]. Though being highly accurate, it is not a real-time SLR system. This issue is dealt with in this paper. In spite of the dataset being small, our system has achieved an average confidence rate of 85.45%.

## 6  Conclusion and Future Works

Sign languages are kinds of visual languages that employ movements of hands, body, and facial expression as a means of communication. Sign languages are important for specially-abled people to have a means of communication. Through it, they can communicate and express and share their feelings with others. The drawback is that not everyone possesses the knowledge of sign languages which limits communication. This limitation can be overcome by the use of automated Sign Language Recognition systems which will be able to easily translate the sign language gestures into commonly spoken language. In this paper, it has been done by TensorFlow object detection API. The system has been trained on the Indian Sign Language alphabet dataset. The system detects sign language in real-time. For data acquisition, images have been captured by a webcam using Python and OpenCV which makes the cost cheaper. The developed system is showing an average confidence rate of 85.45%. Though the system has achieved a high average confidence rate, the dataset it has been trained on is small in size and limited.

In the future, the dataset can be enlarged so that the system can recognize more gestures. The TensorFlow model that has been used can be interchanged with another model as well. The system can be implemented for different sign languages by changing the dataset.